\documentclass[10pt,twocolumn,letterpaper]{article}

\usepackage{cvpr}              
\definecolor{cvprblue}{rgb}{0.21,0.49,0.74}

\def\paperID{*****} 
\def\confName{CVPR}
\def\confYear{2026}

\title{ERGO: Excess-Risk-Guided Optimization for High-Fidelity Monocular 3D Gaussian Splatting}

\begin{document}

\author{
Zehua Ma\textsuperscript{1}\thanks{Equal contribution.} \quad
Hanhui Li\textsuperscript{1}\footnotemark[1] \quad
Zhenyu Xie\textsuperscript{3} \quad
Xiaonan Luo\textsuperscript{4} \\
Michael Kampffmeyer\textsuperscript{5} \quad 
Feng Gao\textsuperscript{2}\footnotemark[2] \quad
Xiaodan Liang\textsuperscript{1,3}\thanks{Corresponding author.} \\
\textsuperscript{1}Shenzhen Campus of Sun Yat-sen University \quad
\textsuperscript{2}Peking University \\
\textsuperscript{3}Mohamed bin Zayed University of Artificial Intelligence \\
\textsuperscript{4}Guilin University of Electronic Technology \quad
\textsuperscript{5}UiT The Arctic University of Norway \\
}

\maketitle
\begin{abstract}
Generating 3D content from a single image remains a fundamentally challenging and ill-posed problem due to the inherent absence of geometric and textural information in occluded regions. While state-of-the-art generative models can synthesize auxiliary views to provide additional supervision, these views inevitably contain geometric inconsistencies and textural misalignments that propagate and amplify artifacts during 3D reconstruction. To effectively harness these imperfect supervisory signals, we propose an adaptive optimization framework guided by excess risk decomposition, termed ERGO. Specifically, ERGO decomposes the optimization losses in 3D Gaussian splatting into two components, i.e., excess risk that quantifies the suboptimality gap between current and optimal parameters, and Bayes error that models the irreducible noise inherent in synthesized views. This decomposition enables ERGO to dynamically estimate the view-specific excess risk and adaptively adjust loss weights during optimization. Furthermore, we introduce geometry-aware and texture-aware objectives that complement the excess-risk-derived weighting mechanism, establishing a synergistic global-local optimization paradigm. Consequently, ERGO demonstrates robustness against supervision noise while consistently enhancing both geometric fidelity and textural quality of the reconstructed 3D content. Extensive experiments on the Google Scanned Objects dataset and the OmniObject3D dataset demonstrate the superiority of ERGO over existing state-of-the-art methods.
\end{abstract}    
\section{Introduction}
\label{sec:intro}

\begin{figure*}[!t]
  \centering
  \includegraphics[width=1.0\hsize]{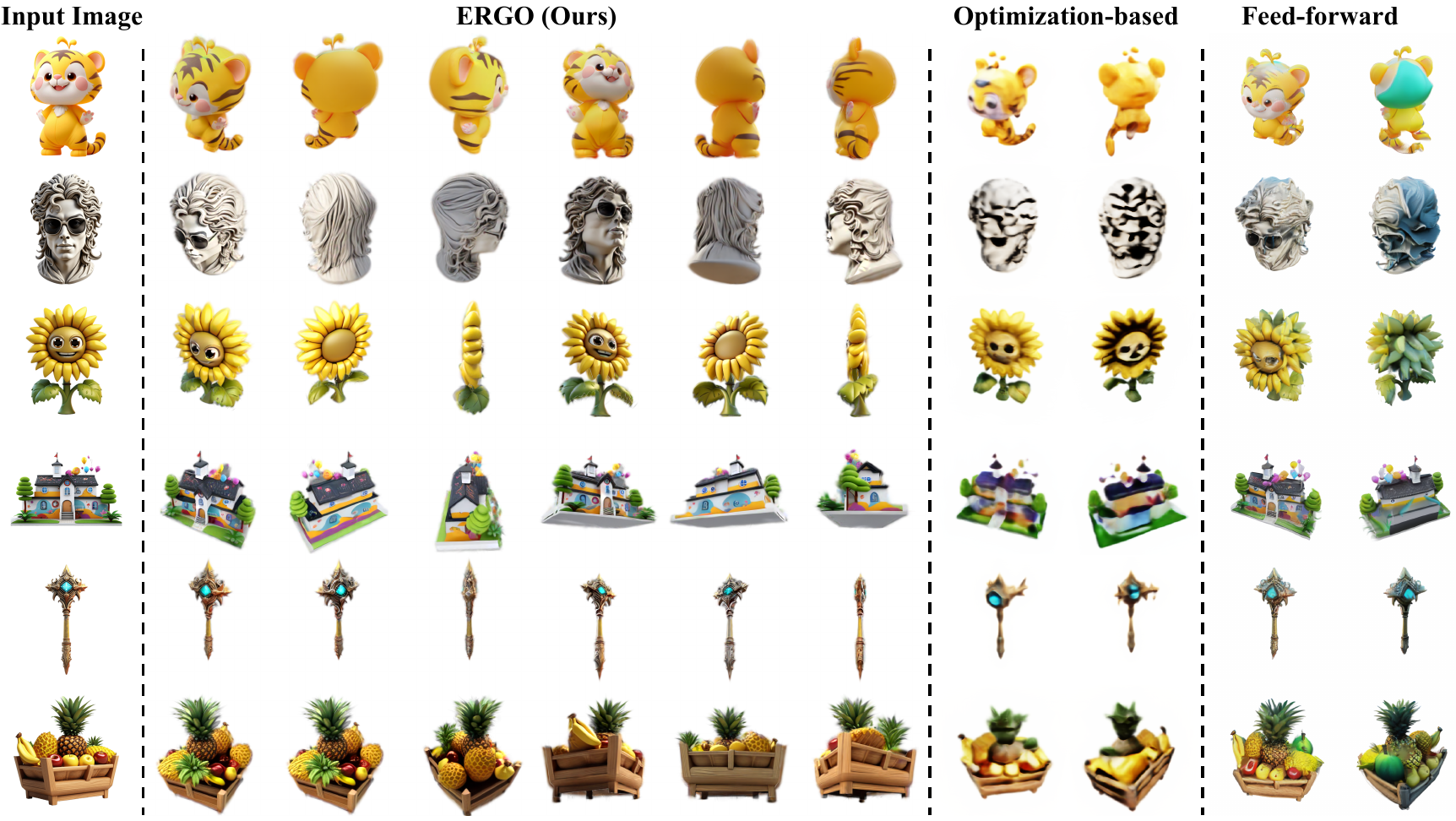}
  \caption{Given single-view images as inputs, the proposed ERGO method can generate 3D objects with better texture consistency and fidelity, compared with state-of-the-art optimization-based methods (e.g., \cite{liu2024syncdreamer}) and feed-forward large models (e.g., \cite{openlrm}).}
  \label{fig: teaser}
\end{figure*}

Single-image 3D content generation (as illustrated in Figure~\ref{fig: teaser}) has attracted considerable attention and facilitated a wide range of applications spanning virtual reality, augmented reality, and game development \cite{wang2025diffusion}. Nevertheless, ensuring cross-view texture consistency and fidelity of the generated 3D content remains challenging due to the limited information provided by single-view images. Existing methods for this task can be broadly categorized into two mainstream paradigms, namely, feed-forward large reconstruction models (LRMs) and optimization-based methods. Specifically, LRMs \cite{jun2023shape, hong2024lrm, li2024instant3d} leverage large-scale 3D datasets to train end-to-end models for direct 3D content prediction. These models exhibit high inference efficiency, yet they demand substantial computational resources for training and often suffer from degraded performance on out-of-distribution data. In contrast, optimization-based methods ~\cite{tang2023makeit3d, Magic123, sun2023dreamcraft3d} typically consist of an iterative optimization process driven by heuristic objectives, such as score distillation sampling (SDS) \cite{poole2022dreamfusion} that exploit texture priors from pretrained image generative models. However, these methods are prone to generating results with cross-view inconsistency and blurry textures, due to inconsistent generative priors induced by diverse camera poses.

One potential solution to mitigate the information scarcity in unobserved views is to employ multi-view diffusion (MVD) models \cite{liu2023zero1to3, shi2023zero123plus, shi2023MVDream, wang2023imagedream, long2024wonder3d} to generate complementary views. With these synthesized views and cutting-edge inverse rendering techniques, e.g., 3D Gaussian splatting (3DGS) \cite{kerbl3Dgaussians,wu2024recent}, conducting an efficient optimization pipeline becomes feasible. This strategy further has the potential to mitigate the blurry textures caused by SDS-based optimization methods, as generated content is grounded by the explicit geometry representation of 3DGS.  

\begin{figure*}[t]
  \centering
  \includegraphics[width=1.0\hsize]{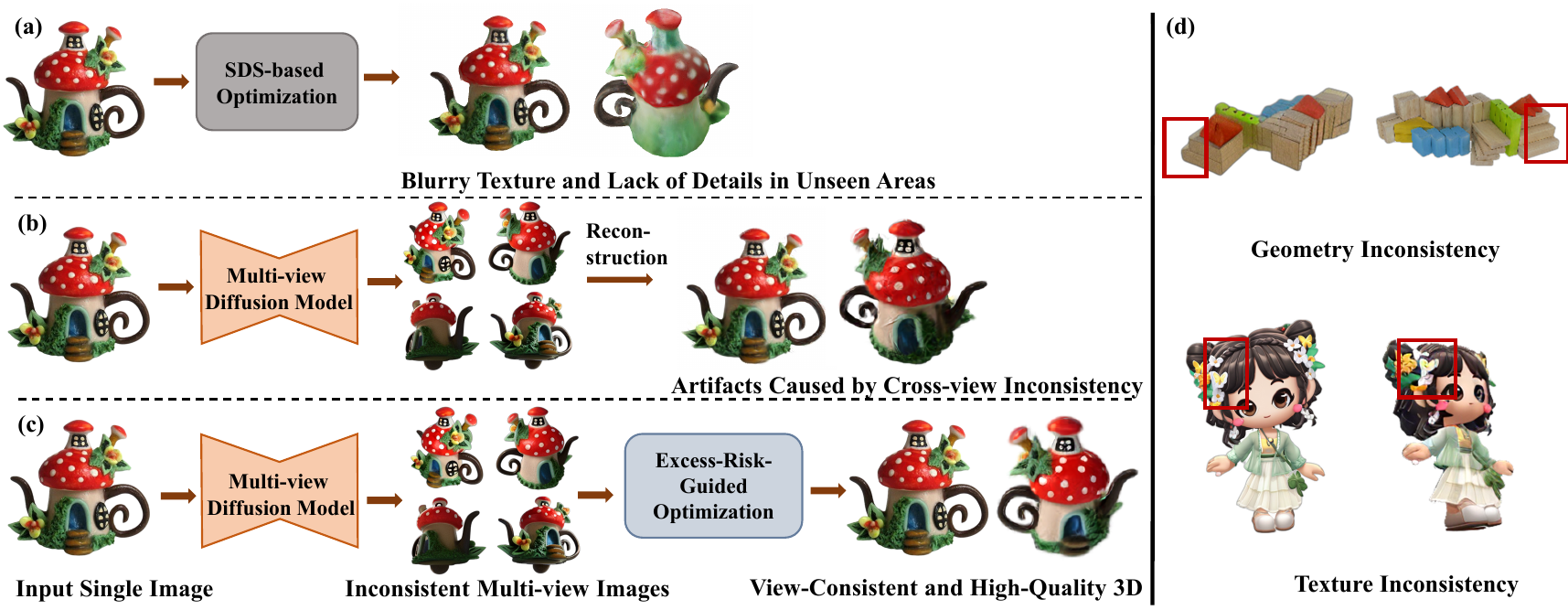}
  \caption{Comparison of various optimization paradigms for single-image-to-3D generation, including (a) optimization-based methods, (b) direct reconstruction with multi-view synthesized images, and (c) the proposed ERGO framework with adaptive objective weights.  (d) Illustration of two types of inconsistency caused by the direct reconstruction with multi-view inconsistent images.}
  \label{fig: motivation}
\end{figure*}

Nevertheless, a direct integration of MVD models and optimization-based methods tends to introduce new types of artifacts due to inherent inconsistencies across the generated multi-view images. As shown in Figure~\ref{fig: motivation}, these inconsistencies can be geometric or textural: geometry inconsistencies are mainly caused by the imperfect 3D modeling ability of existing MVD models, while texture inconsistencies result from the decoupling of texture generation from explicit geometric constraints. Consequently, such view-inconsistent images introduce spurious supervision signals, thereby hindering the stability and convergence of the optimization process.

To address the consistency limitations inherent in current multi-view optimization frameworks, we propose an excess-risk-guided optimization (ERGO) framework in this paper. Unlike conventional optimization-based approaches that assign heuristic and uniform weights to loss objectives across all views, ERGO adaptively estimates and adjusts the weights of each objective during the iterative optimization process. Specifically, ERGO stems from excess risk decomposition \cite{tengtowards, oneto2025informed, he2024robust} that decomposes empirical risk (optimization error) into excess risk and Bayes error. Excess risk measures the discrepancy between current model parameters and theoretical optimal parameters, reflecting the potential of the model for further improvement via optimization. In contrast, Bayes error is inherently attributed to noise in the supervision signals, which corresponds to the geometric and textural inconsistencies in our context. Accordingly, estimating excess risk allows us to dynamically modulate the optimization process and mitigate the adverse effects of Bayes error. We achieve this by adaptively assigning higher global weights to loss objectives and views with greater excess risk, as these components are more informative for guiding the model toward the optimal parameter space.

Furthermore, to complement the global adaptive weighting from excess risk estimation, we introduce a geometry-aware objective and a texture-aware objective. The geometry-aware objective focuses on local geometric consistency across multi-view images. It leverages visibility maps generated via 3DGS to adaptively adjust the loss weight of each local region based on its geometric reliability. Meanwhile, the texture-aware objective models regional texture complexity to facilitate local texture fidelity and detail preservation. This global-local adaptive design enables ERGO to simultaneously achieve inter-view consistency (via global modulation) and intra-view fine-grained quality (via local adjustment), effectively alleviating the geometric and textural artifacts arising from naive MVD-optimization integration.

Overall, our contributions can be summarized as follows:
\begin{itemize}
    \item We propose ERGO, the first single-image to 3D framework that enables adaptive multi-view optimization via excess risk decomposition;
    \item A geometry-aware objective and a texture-aware objective are introduced to perform local adaptive modulation;
    \item Extensive experiments on two public benchmarks (i.e., GSO~\cite{downs2022google}, OmniObject3D ~\cite{wu2023omniobject3d}) demonstrate that ERGO outperforms state-of-the-art methods both qualitatively and quantitatively.
\end{itemize}

\section{Related Work}
\label{sec:relatedwork}

\subsection{Optimization-based Methods}
With the success of image diffusion models, many optimization-based methods have been proposed for novel-view synthesis and 3D content creation. DreamFusion~\cite{poole2022dreamfusion} first introduces the score distillation sampling (SDS) strategy, which enables the generation of plausible 3D content by leveraging image diffusion models to mitigate the need for 3D datasets. A variational improvement of SDS is introduced in \cite{wang2024prolificdreamer}, which reduces texture oversaturation and improves diversity. To improve content fidelity, some researchers~\cite{melas2023realfusion, Magic123} further explore textual inversion, which was originally designed for image customization.

More recently, a series of studies~\cite{lin2023magic3d, Magic123, tang2023makeit3d, chen2023fantasia3d} adopt multi-stage optimization strategies to achieve high-resolution and detail-rich 3D models. For example, Magic3D \cite{lin2023magic3d} and Magic123 \cite{Magic123} optimize low-resolution neural radiance fields (NeRFs) in their first stage. In the second stage, they transfer NeRFs to a more efficient 3D representation \cite{shen2021dmtet} to generate high-resolution meshes. Fantasia3D \cite{chen2023fantasia3d} and HumanNorm~\cite{huang2024humannorm} disentangle geometry and texture, recovering finer geometries and achieving photorealistic rendering. To further enhance the controllability of 3D generation, NeuralLift~\cite{xu2023neurallift}, 3DStyle-Diffusion~\cite{yang20233dstyle}, and Control3d~\cite{chen2023control3d} incorporate additional information (e.g., depth and sketch) for optimization. 

While the above methods demonstrate superior performance in zero-shot 3D synthesis, they still suffer from the Janus (multi-faced) problem and blurred results caused by their insufficient perception of 3D geometry. Moreover, they typically require minutes or even hours to generate a single 3D model. To address these limitations, our proposed ERGO framework employs 3D Gaussians to achieve efficient modeling of geometry and provide geometric priors, enabling the generation of realistic textures.

\subsection{Feed-Forward Models}
The advance of network architectures and large-scale 3D datasets facilitates data-driven feed-forward models~\cite{chen2023single, cheng2023sdfusion, zhang20233dshape2vecset, xue2025gen, wang2025geco, chen2025sam}, alleviating the prohibitive computational cost of optimization-based methods. For example, Point·E~\cite{nichol2022point} employs a transformer-based model to regress RGB point clouds from input images and LION~\cite{vahdat2022lion} constructs a hierarchical VAE, encoding point clouds into a latent space for reconstruction. Both of these works generate explicit 3D assets. In contrast, Shap-E~\cite{jun2023shape} generates parameters for implicit functions that can be rendered as both textured meshes and neural radiance fields. These early research efforts produce plausible 3D models with simple geometry and texture, yet they yield collapsed results under complex conditions due to the lack of large-scale data. Recently, by utilizing large-scale 3D datasets~\cite{objaverse,objaverseXL,wu2023omniobject3d} for training, the generality and quality of 3D generative models have been improved significantly.

As in feed-forward methods, LRM~\cite{openlrm} is a pioneer in utilizing a regression model to predict NeRF representations from a single image, with subsequent works expanding it to image-to-gaussian and image-to-mesh generation. \cite{zou2024triplane} employs a hybrid triplane-based Gaussian representation, effectively balancing rendering speed and quality. Moving away from regression-based models, ~\cite{NEURIPS2023_ea1a7f7b, zhang2024clay, wu2024direct3d, zhao2025hunyuan3d} incorporate latent diffusion transformers (DiTs) for iterative denoising and learning 3D distributions in latent space. Clay~\cite{zhang2024clay} further scales up its generative model, enhancing its generalization and diversification capabilities. To tackle the limitations of implicit latent representations in efficient encoding, Direct3D~\cite{wu2024direct3d} adopts explicit triplane latent representations and employs a geometric mapping network to predict 3D occupancy grids.

Given a single reference image, these feed-forward models tend to produce blurry textures and overly smooth geometries when applied to unseen viewpoints. An alternative approach involves combining multi-view generation models to supplement information in unseen viewpoints before executing the 3D reconstruction process. Building upon the LRM architecture, Instant3D~\cite{li2024instant3d} and InstantMesh~\cite{xu2024instantmesh} address the Janus problem and generate high-fidelity 3D meshes. However, maintaining multi-view consistency remains challenging and often leads to low-quality geometric structures. Craftsman~\cite{li2024craftsman} employs a 3D diffusion model conditioned on multiple views to generate coarse 3D geometry and a normal-based geometry refiner to enhance surface details significantly. LGM~\cite{tang2024lgm} further improves upon this by utilizing Gaussian splatting for higher fidelity 3D results. Additionally, ~\cite{szymanowicz2024splatter, charatan2024pixelsplat, chen2024mvsplat} parameterizes 3D assets using pixel-aligned Gaussians and fuses multiple per-view Gaussians, enabling the efficient generation of high-resolution 3D content. While these models achieve high inference efficiency, they typically incur significant training costs and exhibit limited generalization to out-of-distribution scenarios. ERGO, by contrast, formulates 3D reconstruction as an optimization problem and directly exploits the rich priors of pretrained image generative models, effectively mitigating these limitations.

\subsection{Multi-view Diffusion Model}
To address view ambiguities, researchers have used multi-view diffusion~\cite{shi2023MVDream, zhou2024multi3d, Yang_2024_CVPR, wen2025ouroboros3d, cai2025baking, zhang2025ar} to supplement missing information. Pretrained on a billion-scale dataset of image-text pairs, image generation models~\cite{chen2023pixart, esser2024scaling, batifol2025flux} have gained abundant 3D prior knowledge. By simply fine-tuning the image diffusion model on 3D datasets, Zero1-to-3~\cite{liu2023zero1to3} is able to generate novel views of the same object when given a single view and the corresponding camera transformations. However, the multi-view images produced by Zero1-to-3 often exhibit inconsistencies due to the lack of information interaction across the different views. 
A series of studies~\cite{hollein2024viewdiff, deng2023mv, tang2024mvdiffusion++, huang2023epidiff, huang2025mv} have been dedicated to improving the consistency of multi-view images, which is crucial for subsequent 3D reconstruction. 
Zero123++~\cite{shi2023zero123plus} combines six images in a $3\times2$ layout into a single frame, generating multiple views in one forward pass. SyncDreamer~\cite{liu2024syncdreamer} utilizes a 3D feature volume as a unified feature representation for multi-view images, facilitating information fusion across different views. Wonder3D~\cite{long2024wonder3d} learns the joint distribution of normal maps and RGB images, enhancing the alignment of 3D object geometry and texture.

Certain studies have turned their attention to video generation models~\cite{blattmann2023align, blattmann2023stable, chai2023stablevideo}, given that video models' inherent temporal and spatial coherence dovetail perfectly with the consistency required for multi-view generation. Various 3D guidance and conditions have been incorporated into video generation models to improve multi-view consistency. For instance, VideoMV~\cite{zuo2024videomv} adopts a 3D-aware denoise sampling strategy that effectively inserts images rendered from 3D into its denoising process. Hi3D~\cite{yang2024hi3d} leverages depths as conditions for a video-to-video refiner, generating high-resolution views with detailed texture. RECONX~\cite{liu2024reconx} builds a global point cloud and uses it as 3D structure guidance. Additionally, subsequent research has improved upon model flexibility, allowing for more freedom in inputs and outputs. Vivid-1-to-3~\cite{kwak2024vivid} combines diffusion models for both image and video to generate video frames conditioned on camera trajectories. V3D~\cite{peng2010v3d} incorporates a pixelNeRF encoder, which could be seamlessly adapted to any number of input images. SV3D~\cite{voleti2024sv3d} further adds explicit camera controls for novel view synthesis so that azimuths can be irregularly spaced and elevations can vary per view. However, the multi-view images generated by these methods still suffer from cross-view inconsistencies. Our ERGO framework addresses this issue through excess-risk-guided optimization, enabling the generation of 3D reconstructions with strong geometric consistency and realistic textures.
\section{Preliminaries}

\begin{figure*}[!t]
  \centering
  \includegraphics[width=1.0\hsize]{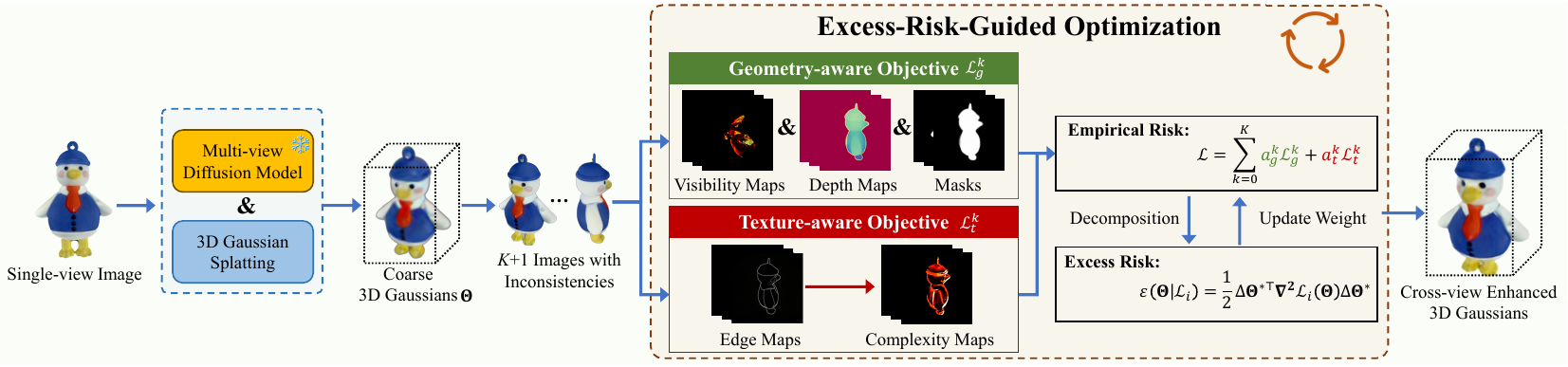}
  \caption{The proposed ERGO framework for single-image 3D content generation. Given coarse 3D Gaussians and synthesized images with inconsistencies, ERGO not only estimates excess risk to mitigate inconsistencies and modulates iterative optimization globally, but also leverages the geometry-aware objective and the texture-aware objective to achieve localized refinement.}
  \label{fig: framework}
\end{figure*}

\subsection{3D Gaussian Splatting} 
3D Gaussian Splatting (3DGS) \cite{kerbl3Dgaussians} is based on an explicit representation that uses a set of points with optimizable attributes for rendering an object or scene, e.g., position, color, and density. The rendering process is conducted by projecting each Gaussian point onto an image plane, which can be formulated as,
\begin{equation}
    C=\sum_{j \in N} c_j \alpha_j \prod_{k=1}^{j-1}\left(1-\alpha_k\right).
\end{equation}
Here $c_j$ and $\alpha_j$ denote the color and the opacity of each point, respectively, $N$ is the number of points, and $C$ represents the color of a pixel. The Gaussian attributes can be predicted by a pretrained model or optimized from scratch when sufficient multi-view images are available.

\subsection{Optimization-based 3D Generation}
We adopt the optimization-based paradigm for image-to-3D generation, owing to its flexibility in model selection and potential for targeted enhancement. A commonly used strategy is score distillation sampling (SDS) \cite{poole2022dreamfusion, liu2023zero1to3}, which minimizes the difference between the noise of a reference image $I_r$ and that of a rendered image $I$:
\begin{equation}
\mathcal{L}_{\mathrm{SDS}}(I,I_r)=\mathbb{E}_{t, p, \epsilon}\left[w(t)\left(\epsilon_\phi\left(I; t, I_r, p\right)-\epsilon\right) \frac{\partial I}{\partial \bf{\Theta}}\right].
\label{eq: SDS}
\end{equation}
Here, $t$ is a randomly sampled timestep and $p$ is a randomly sampled camera pose. $\epsilon$ denotes Gaussian noise sampled from a standard normal distribution, $w(t)$ is a timestep-related weight, $\epsilon_\phi(\cdot)$ denotes a pretrained noise predictor, and $\bf{\Theta}$ denotes parameters to be optimized.

\section{Methodology}

In this section, we present the details of our excess-risk-guided optimization framework (ERGO) for monocular 3D Gaussian splatting (3DGS). The core of ERGO is to formulate 3DGS within a weighted multi-objective paradigm (Sec. \ref{sec: overall_framework}), where the weights assigned to individual objectives are conditioned on the excess risks estimated from noisy auxiliary images. The major objectives of ERGO include a geometry-aware objective (Sec. \ref{sec: GAO}) and a texture-aware objective (Sec. \ref{sec: TAO}), which are specifically devised to emphasize image regions with geometric consistency and rich details, respectively. Detailed strategies for excess risk estimation and weight updates are presented in Sec. \ref{sec: optimization}.

\subsection{Overall Framework}
\label{sec: overall_framework}
Although it is intuitive to leverage images generated by an MVD model to optimize Gaussian attributes, a naive strategy that directly optimizes over the generated images will inevitably lead to geometric and textural inconsistencies. Such inconsistencies arise from the lack of explicit 3D modeling in most existing MVD models and distribution discrepancies between training and test data. Therefore, we propose the ERGO framework to adaptively adjust the weights of the optimization objectives according to the quality of the generated images and their effect on the optimization process, as shown in Figure \ref{fig: framework}.

Specifically, given a reference image $I_0$ and a set of $K$ auxiliary images $\mathcal{A} = \{I_1,..., I_K\}$ generated by an off-the-shelf MVD model (e.g., \cite{shi2023zero123plus}), the ERGO framework is designed to optimize Gaussian attributes $\bm{\Theta}$ to solve the following multi-objective optimization problem:
\begin{equation}
    \mathop {\min }\limits_{\bf{\Theta}}\sum\limits_{m = 1}^M {\sum\limits_{k = 0}^{K} {{a_{mk}}{\mathcal{L}_{mk}}({\bf{\Theta}}, {I_k})}},
    \label{eq: multi_object}
\end{equation}
where $M$ denotes the number of objectives, $a_{mk} \in [0, 1)$ represents the weight of ${\mathcal{L}_{mk}}$, and $I_k \in \{I_0,  \mathcal{A}\}$. We assume ${\bf{a}} = \{a_{mk}\}$ lies in a probability simplex, and hence $\sum a_{mk} =1$. If each ${\mathcal{L}_{mk}}$ satisfies certain constraints (e.g., Lipschitz continuity) \cite{he2024robust}, Eq. (\ref{eq: multi_object}) can achieve a Pareto stationary solution, i.e., an optimal solution that has no feasible direction to reduce any of the weighted loss terms ${a_{mk}}{\mathcal{L}_{mk}}({\bf{\Theta}}, {I_k})$ without increasing others. While this condition is non-trivial to satisfy, as some objectives are non-convex or their tight upper bounds are difficult to derive, our empirical results demonstrate that the proposed method outperforms the baseline model without adaptive weights. 

To model and suppress per-view noise in $\mathcal{A}$, we propose to decompose the expected loss of ${\bf{\Theta}}$ with respect to ${\mathcal{L}_{mk}}$ into the excess risk and the Bayes error as
\begin{equation}
    R({\bf{\Theta}}|\mathcal{L}_{mk}) = \varepsilon({\bf{\Theta}}|\mathcal{L}_{mk}) + \varepsilon({\bf{\Theta}}^{*}|\mathcal{L}_{mk}),
    \label{eq: excess_risk}
\end{equation}
where $R({\bf{\Theta}}|\mathcal{L}_{mk})=\mathbb{E}[\mathcal{L}_{mk}(\bf{\Theta})]$ denotes the expected loss and $\varepsilon({\bf{\Theta}}|\mathcal{L}_{mk})$ represents the excess risk. ${\bf{\Theta}}^{*}$ denotes the ideal but unknown parameters. The Bayes error $\varepsilon({\bf{\Theta}}^{*}|\mathcal{L}_{mk})$ is irreducible, due to the stochastic nature of MVD generation and the partial observability of ground-truth multi-view data. Fortunately, we can approximate $\varepsilon({\bf{\Theta}}|\mathcal{L}_{mk})$ using the gradients of the empirical risk of ${\bf{\Theta}}$ and its Taylor expansion, without explicitly estimating ${\bf{\Theta}}^{*}$ or $\varepsilon({\bf{\Theta}}^{*}|\mathcal{L}_{mk})$, as detailed in Sec. \ref{sec: optimization}.

By rearranging Eq. (\ref{eq: excess_risk}) as $\varepsilon({\bf{\Theta}}|\mathcal{L}_{mk}) = R({\bf{\Theta}}|\mathcal{L}_{mk}) - \varepsilon({\bf{\Theta}}^{*}|\mathcal{L}_{mk})$, it is evident that the excess risk quantifies the gap between the performance of the current parameters and that of the optimal parameters. Therefore, we can use the gradient norm of $\varepsilon({\bf{\Theta}}|\mathcal{L}_{mk})$, denoted as $||\nabla\varepsilon({\bf{\Theta}}|\mathcal{L}_{mk})||$, to adaptively adjust $a_{mk}$ and promote convergence. Under the probability simplex assumption of $\bf{a}$, the ERGO framework allows us to approximate a Pareto stationary among the objectives across all views. Note that Eq. (\ref{eq: multi_object}) serves as a global weighting mechanism, and hence we introduce the geometry-aware objective and the texture-aware objective in the subsequent sections to achieve localized adjustments of objective weights.

\begin{figure}[!t]
  \centering
  \includegraphics[width=1.0\hsize]{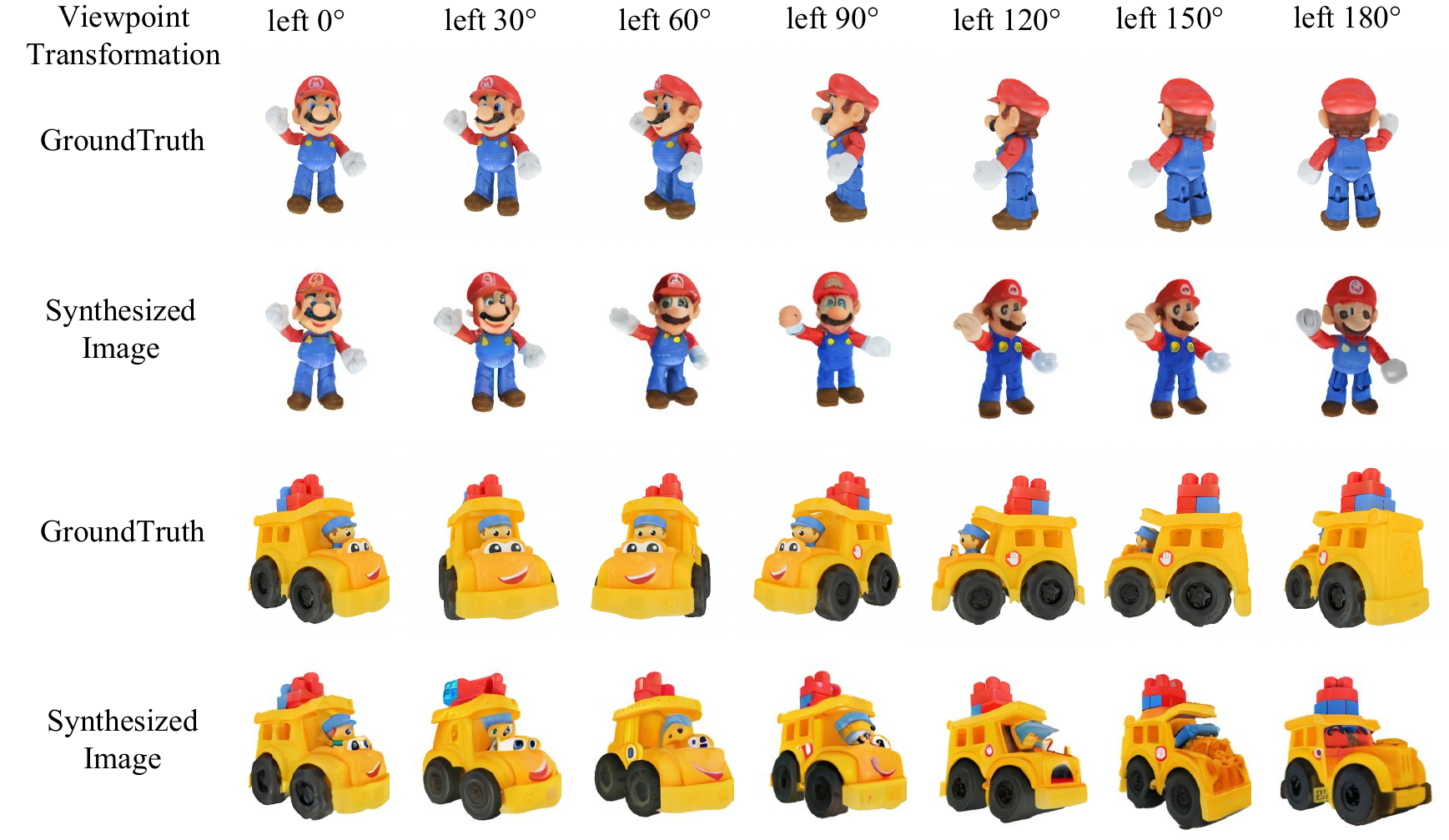}
  \caption{The performance of the MVD baseline degrades as the viewpoint transformation magnitude increases.}
  \label{fig: view_degrade}
\end{figure}

\subsection{Geometry-Aware Objective}
\label{sec: GAO}
We introduce the geometry-aware objective to alleviate the absence of geometric constraints in conventional optimization objectives of MVD models (e.g., SDS). The geometry-aware objective consists of two components: (i) explicit geometric loss terms, and (ii) an adaptive loss term modulated by geometric properties.

\textbf{Geometry Correction}. We first propose a geometry correction step to construct a coarse 3D Gaussian model $G_c$, which serves as pseudo-geometric ground truth. We follow \cite{liu2023zero1to3} for constructing $G_c$ by minimizing the SDS loss over $\mathcal{I} = \{{I}_0, \mathcal{A}\}$. We then use $G_c$ to calculate depth maps $\mathcal{D} = \{D_0, ..., D_K\}$ and object masks $\mathcal{M} = \{M_0, ..., M_K\}$ corresponding to each view in $\mathcal{I}$. Each depth map $D_k$ is normalized to $[0, 1]$, and the elements belonging to the object in $M_k$ are 1, otherwise 0. Furthermore, to mitigate geometric inconsistency across $\mathcal{I}$, rather than using $\mathcal{A}$ for subsequent optimizations, we also use $G_c$ to render the counterparts of the images in $\mathcal{A}$ and employ DDIM inversion \cite{mokady2022nulltextinversioneditingreal} to convert them back into the noise latent space of the MVD model for denoising. This allows us to acquire auxiliary images with refined textures and geometric consistency. For conciseness, we retain the notation $\mathcal{A}$ for these refined auxiliary images.

Our geometry-aware objective ${\mathcal{L}}_g$ is formulated as follows:
\begin{equation}
    {\mathcal{L}}_{g} = \lambda_{v}{{{\mathcal{L}}_{v}}} + \lambda_{d}{{{\mathcal{L}}_{d}}} + \lambda_{m}{{{\mathcal{L}}_{m}}}.
    \label{eq: gao}
\end{equation}
$\lambda_{v}$, $\lambda_{d}$, and $\lambda_{m}$ are scalar coefficients to normalize the loss terms and preserve the probability simplex property of the adaptive weights $\bf{a}$. ${{{\mathcal{L}}_{d}}}$ and ${{{\mathcal{L}}_{m}}}$ are two explicit geometric losses defined as ${{{\mathcal{L}}_{d}}}=\sum\nolimits_{k = 0}^{K}\left\|D_k'-D_k \right\|_2^2$ and ${{{\mathcal{L}}_{m}}}=\sum\nolimits_{k = 0}^{K}\left\|M_k'-M_k \right\|_2^2$, where $D_k'$ and $M_k'$ are the depth map and object mask predicted during the optimization process, respectively. ${{{\mathcal{L}}_{v}}}$ is the adaptive loss term modulated by visibility maps, which is the core of our geometry-aware objective. The key idea of $\lambda_{v}$ is to assign higher weights to visible image regions, as these regions are geometrically more reliable, as shown in Figure \ref{fig: view_degrade}.

\begin{figure}[!t]
  \centering
  \includegraphics[width=1.0\hsize]{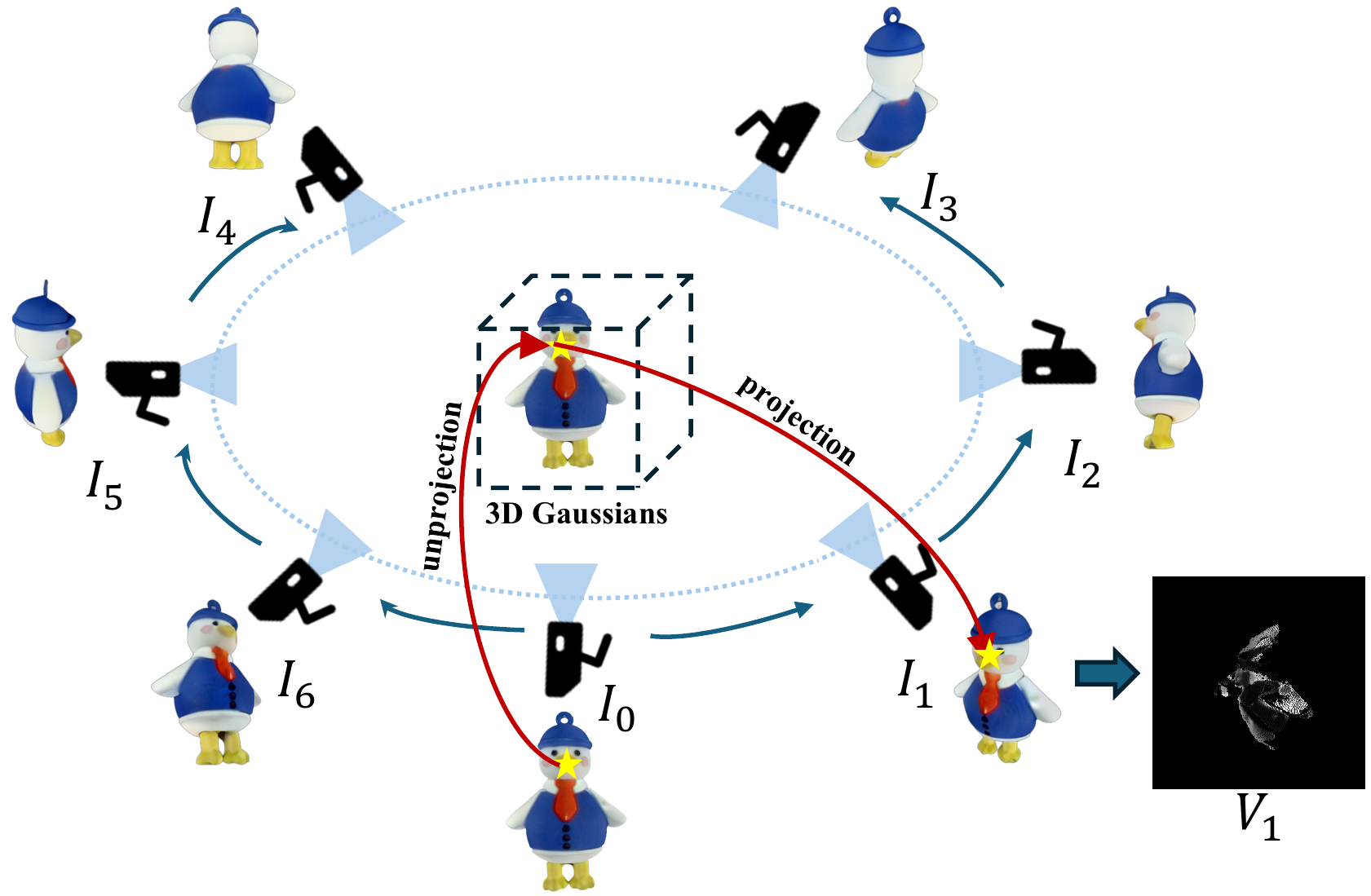}
  \caption{Illustration of visibility map generation. We lift the pixels from one image into 3D space and project them onto the adjacent image to identify the corresponding pixels. We then calculate the differences between these corresponding pixels to generate the visibility map.}
  \label{fig: overlap}
\end{figure}

\begin{figure*}[!t]
  \centering
  \includegraphics[width=1.0\hsize]{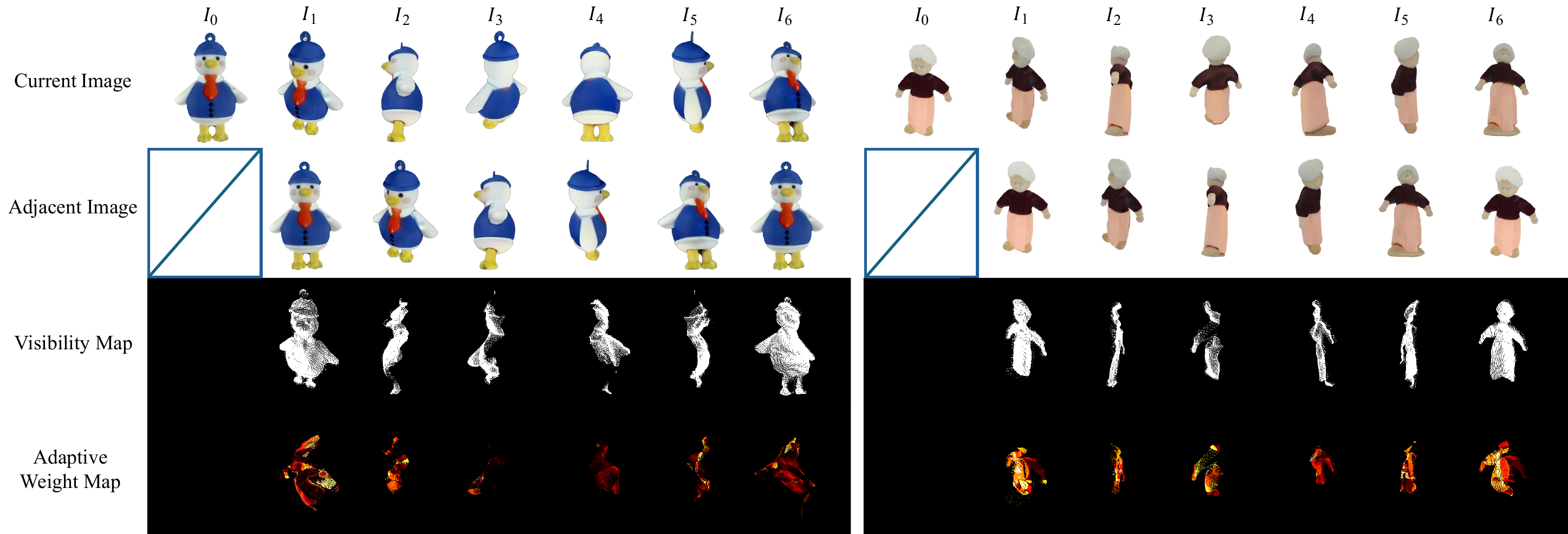}
  \caption{Visualizations of visibility maps and adaptive weight maps conditioned on discrepancy. These weight maps have high values in visible and consistent regions.}
  \label{fig: visibility}
\end{figure*}

Due to the coarse geometry of $G_c$, we refrain from rendering methods (e.g., z-buffering) for visibility map computation. As illustrated in Figure~\ref{fig: overlap}, we leverage $\mathcal{D}$ and $\mathcal{M}$ to establish pixel-wise correspondences between adjacent views and compute visibility maps. Let ${\bf{p}}_k$ denote the camera parameters for the $k$-view. We estimate the pixel-wise correspondences between $D_k$ and $D_{k+1}$ by first lifting the pixels in $D_k$ to the 3D space with the inverse of ${\bf{p}}_k$, followed by projecting them to the image plane of view $k+1$ using $p_{k+1}$. This yields a warped depth map $D_{k \to k+1}$, representing the depth values of points from view $k$ as observed in view $k+1$. The visibility map $V_{k+1}$ for view $k+1$ is then defined as,
\begin{equation}
    V_{k+1} = \max(0, M_{k+1} \odot(D_{k \to {k+1}} - D_{k+1})),
    \label{eq: vis_map}
\end{equation}
where $\odot$ denotes element-wise multiplication. The intuition behind Eq. (\ref{eq: vis_map}) is that if the depth value of a pixel in $D_k$ after the transformation is smaller than that of the corresponding pixel in $D_{k+1}$, then the 3D surface point corresponding to the pixel is nearer the camera and thus visible. 

Leveraging the visibility maps, we formally define the visibility-aware loss term ${\mathcal{L}}_{v}$ as,
\begin{equation}
    {\mathcal{L}}_{v} = \sum_{k=0}^K V_k \odot W_i \odot\left\| I_k' - I_k\right\|_2^2 + (1-V_k)\odot\left\| I_k' - I_k\right\|_2^2,
    \label{eq: loss_vis}
\end{equation}
where the first and second terms on the right-hand side respectively modulate the supervision weights assigned to the visible and invisible pixel regions. $W_i = 1 - (I_{k-1 \to k} - I_k)^s$ is an adaptive weight that is conditioned on the discrepancy between $I_k$ and $I_{k-1 \to k}$, where the latter is obtained via the same warping process as $D_{k-1 \to k}$. $s > 0$ is a hyperparameter. $W_i$ assigns higher weights to regions where the warped content $I_{k-1 \to k}$ closely matches $I_k$, indicating consistent multi-view geometry for supervision. We set $W_0 = V_0 = \bf{1}$. Examples of the visibility maps and weight maps are visualized in Figure \ref{fig: visibility}.

\subsection{Texture-Aware Objective}
\label{sec: TAO}

In addition to the geometry-aware objective, we also introduce a texture-aware objective to better exploit the texture priors in the MVD base model. However, the conventional SDS loss tends to over-smooth fine-grained textures \cite{katzirnoise}, which is exacerbated when learning on view-inconsistent images. To address this issue and facilitate the learning of texture details, our texture-aware objective adaptively assigns SDS loss weights to distinct image regions based on their texture complexity. 

Formally, our texture-aware objective $\mathcal{L}_t$ is defined as,
\begin{equation}
    \mathcal{L}_t = \lambda_t\sum_{k=0}^K (1 - d({\bf{c}}, {\bf{c}}_k)) W_k^c \odot \mathcal{L}_{\mathrm{SDS}}(I', I_k).
    \label{eq: tex_obj}
\end{equation}
Here, $\mathcal{L}_{\mathrm{SDS}}$ is the SDS loss defined in Eq. (\ref{eq: SDS}). $\lambda_t$ is a hyperparameter that balances the contribution of $\mathcal{L}_t$ among all objectives. $d({\bf{c}}, {\bf{c}}_k)$ denotes the distance (e.g., Euclidean distance) between a randomly sampled camera center $c$ and the camera center of the reference image $I_k$. The distance is normalized across all $K + 1$ reference images, i.e., $d({\bf{c}}, {\bf{c}}_k):=\frac{d({\bf{c}}, {\bf{c}}_k)}{\sum\nolimits_{j=0}^{K}d({\bf{c}}, {\bf{c}}_j)}$, ensuring reference images closer $\bf{c}$ exert greater influence. $W_k^c$ is a texture complexity weight that adaptively adjusts the SDS loss weight assigned to $I_k$. Various methods can be used to implement $W_k^c$, such as pattern entropy \cite{ke2016texture} and weighted fusion of multiple visual cues \cite{corchs2016predicting}. In this paper, we adopt a lightweight implementation by modeling texture complexity using the pixel-wise edge map of the predicted image $I_k'$. This is achieved efficiently via the classic Sobel operator \cite{gonzalez2009digital}, which can be formulated as,
\begin{equation}
    W_k^c = 1 + Sobel(I_k') \circledast \mathcal{F},
    \label{eq: tex_comp}
\end{equation}
where $\mathcal{F}$ denotes a Gaussian filter employed to smooth out pixel-wise edge responses and $\circledast$ represents the convolution operation. In this way, our texture-aware objective not only prioritizes reference images that are spatially closer to the sampled camera to enhance geometric consistency, but also allocates higher SDS loss weights to texture-rich regions, effectively preserving fine-grained details.

\subsection{Optimization}
\label{sec: optimization}
By integrating the geometry-aware objective and the texture-aware objective, the overall optimization objective of our ERGO framework is formally defined as follows,
\begin{equation}
    \mathcal{L} = \sum_{k=0}^K a_g^k \mathcal{L}_g^k + a^k_t\mathcal{L}^k_t, \quad s.t. \sum_{k=0}^K a_g^k + a^k_t = 1,
    \label{eq: all_obj}
\end{equation}
where the superscript $k$ indexes the components of both adaptive weights and loss objectives corresponding to the $k$-th view. We optimize the Gaussian parameters ${\bf{\Theta}}$ and the adaptive weights $\bf{a}$ alternatively, so that we can analyze each loss term in $\mathcal{L}$ when we update $\bf{a}$ with fixed ${\bf{\Theta}}$. 

Formally, let ${\mathcal{L}}_i \in \{{\mathcal{L}}_g^0,..., {\mathcal{L}}_g^K, {\mathcal{L}}_t^0,..., {\mathcal{L}}_t^K\}$ denote an arbitrary loss component in the overall objective $\mathcal{L}$. To estimate the excess risk of ${\mathcal{L}}_i$, we follow \cite{he2024robust} to assume the optimal Gaussian parameters ${\bf{\Theta}}^{*}$ can be locally approximated as ${\bf{\Theta}}^{*} = {\bf{\Theta}} + \Delta{\bf{\Theta}}^*$, where $\Delta{\bf{\Theta}}^*$ denotes a small perturbation in the local neighborhood of ${\bf{\Theta}}$. Furthermore, the second-order Taylor expansion of ${\mathcal{L}}_i({\bf{\Theta}} + \Delta{\bf{\Theta}})$ is given by,
\begin{equation}
    \mathcal{L}_i(\boldsymbol{\Theta} + \Delta\boldsymbol{\Theta}) \approx 
    \begin{aligned}
        &\mathcal{L}_i(\boldsymbol{\Theta}) + \nabla\mathcal{L}_i(\boldsymbol{\Theta})^{\top}\Delta\boldsymbol{\Theta} \\
        & + \frac{1}{2}\Delta\boldsymbol{\Theta}^{\top}\nabla^2\mathcal{L}_i(\boldsymbol{\Theta})\Delta\boldsymbol{\Theta},
    \end{aligned}
    \label{eq: taylor}
\end{equation}
where $\nabla{\mathcal{L}}_i({{\bf{\Theta}}})$ denotes the gradient of ${\mathcal{L}}_i({{\bf{\Theta}}})$ and $\nabla^2{\mathcal{L}}_i({{\bf{\Theta}}})$ is the corresponding Hessian matrix. With the assumption that ${\bf{\Theta}}^{*}$ is locally optimal, $\nabla\mathcal{L}_i(\boldsymbol{\Theta}^*) = \bf{0}$ and hence, 
\begin{equation}
    \nabla\mathcal{L}_i(\boldsymbol{\Theta}) + \nabla^2\mathcal{L}_i(\boldsymbol{\Theta})\Delta{\bf{\Theta}}^*= \bf{0}.
    \label{eq: local_opt}
\end{equation}
Substituting Eq. (\ref{eq: local_opt}) into Eq. (\ref{eq: taylor}), we have:
\begin{equation}
    \mathcal{L}_i(\boldsymbol{\Theta}^*) = \mathcal{L}_i(\boldsymbol{\Theta}) - \frac{1}{2}
\Delta\boldsymbol{\Theta}^{*\top}\nabla^2\mathcal{L}_i(\boldsymbol{\Theta})\Delta{\boldsymbol{\Theta}}^*.
\label{eq: er_est}
\end{equation}
Based on the excess risk decomposition defined in Eq. (\ref{eq: excess_risk}), we can now estimate $\varepsilon({\bf{\Theta}}|\mathcal{L}_i)$ as,
\begin{equation}
    \varepsilon({\bf{\Theta}}|\mathcal{L}_i) = \frac{1}{2}
\Delta\boldsymbol{\Theta}^{*\top}\nabla^2\mathcal{L}_i(\boldsymbol{\Theta})\Delta{\boldsymbol{\Theta}}^*.
\end{equation}
Here $\nabla^2\mathcal{L}_i(\boldsymbol{\Theta})$ can be computed directly at $\bf{\Theta}$, and $\Delta{\boldsymbol{\Theta}}^*$ can be solved by inverting $\nabla^2\mathcal{L}_i(\boldsymbol{\Theta})$ according to Eq. (\ref{eq: local_opt}), i.e., $\Delta{\boldsymbol{\Theta}}^*=-(\nabla^2\mathcal{L}_i(\boldsymbol{\Theta}))^{-1}\nabla\mathcal{L}_i(\boldsymbol{\Theta})$.

Accordingly, we assign higher adaptive weights to loss terms with larger excess risks. This enables us to focus on under-optimized objectives and mitigate Bayes error (i.e., noise in auxiliary views). Following \cite{he2024robust}, we reformulate the objective of Eq. (\ref{eq: all_obj}) as $\mathop {\min }\limits_{\mathbf{\Theta}} \mathop {\max }\limits_{a_i \in \mathbf{a}}a_i\mathcal{L}_i$, and update the adaptive weight $a_i$ iteratively as follows:
\begin{equation}
    a^{n+1}_i = \frac{a^{n}_i\mathrm{exp}(\eta\varepsilon({\bf{\Theta}}|\mathcal{L}_i))}{\sum_{j=0}^{|\bf{a}|}a^{n}_j\mathrm{exp}(\eta\varepsilon({\bf{\Theta}}|\mathcal{L}_j))},
    \label{eq: a_update}
\end{equation}
where $n$ denotes the $n$-th optimization step and $\eta$ is a hyperparameter that controls the sensitivity of weight updates to excess risk estimates.
\section{Experiments}

\begin{table*}[!t]
    \caption{Quantitative comparison with state-of-the-art methods on GSO~\cite{downs2022google} and OmniObject3D~\cite{wu2023omniobject3d}.}
    \def\arraystretch{1}
    \tabcolsep 11 pt
    \centering
    \begin{tabular}{l c c c c c c c c c c}
        \toprule
        \multirow{2}*{Method} & \multirow{2}*{Venue} & \multicolumn{3}{c}{GSO dataset~\cite{downs2022google}}  & & \multicolumn{3}{c}{OmniObject3D dataset~\cite{wu2023omniobject3d}} \\
        \cmidrule{3-5}  \cmidrule{7-9} 
        & & PSNR$\uparrow$ & SSIM$\uparrow$ &  LPIPS$\downarrow$ & & PSNR$\uparrow$ & SSIM$\uparrow$ &  LPIPS$\downarrow$ \\
        \midrule
        SyncDreamer~\cite{liu2024syncdreamer} & ICLR 2024 & 18.11 & 0.8081 & 0.1768 & & 16.80 & 0.7992 & 0.17821 \\
        Wonder3D~\cite{long2024wonder3d}  &  CVPR 2024  & 17.22 & 0.7781 & 0.2146 & & 14.63 & 0.7715 & 0.2245 \\
        DreamGaussian~\cite{tang2024dreamgaussian} & ICLR 2024 & 20.05 & 0.8048 & 0.1764 & & 18.66 & 0.8072 & 0.1768 \\
        \midrule
        LRM~\cite{openlrm} & ICLR 2024 & 18.10 & 0.7802 & 0.1847 & & 17.26 & 0.8021 & 0.1674 \\
        LGM~\cite{tang2024lgm} & ECCV 2024 & 14.78 & 0.7377 & 0.2795 && 13.31 &  0.7403 & 0.2738  \\
        VideoMV~\cite{zuo2024videomv} &  Alibaba 2024 & 21.06 & 0.8364 & 0.1846 & & 18.75 & 0.8243 & 0.1898 \\
        InstantMesh~\cite{xu2024instantmesh} & Tencent 2024 & 18.27 & 0.8066 & 0.1677 & & 16.82 & 0.8139 & 0.1645 \\
        SAR3D~\cite{chen2024sar3d}&  CVPR 2025 & 17.01 & 0.8006 & 0.2104 & & 15.92 & 0.8090 & 0.1937 \\
        \midrule
        \textbf{ERGO} & \textbf{Ours} & \textbf{21.37}  & \textbf{0.8426} & \textbf{0.1609} & & \textbf{20.24} & \textbf{0.8854} & \textbf{0.1422}  \\
        \bottomrule
    \end{tabular}
    \label{tab:quantitative_results}
\end{table*}

\begin{figure*}[!t]
  \centering
  \includegraphics[width=1.0\hsize]{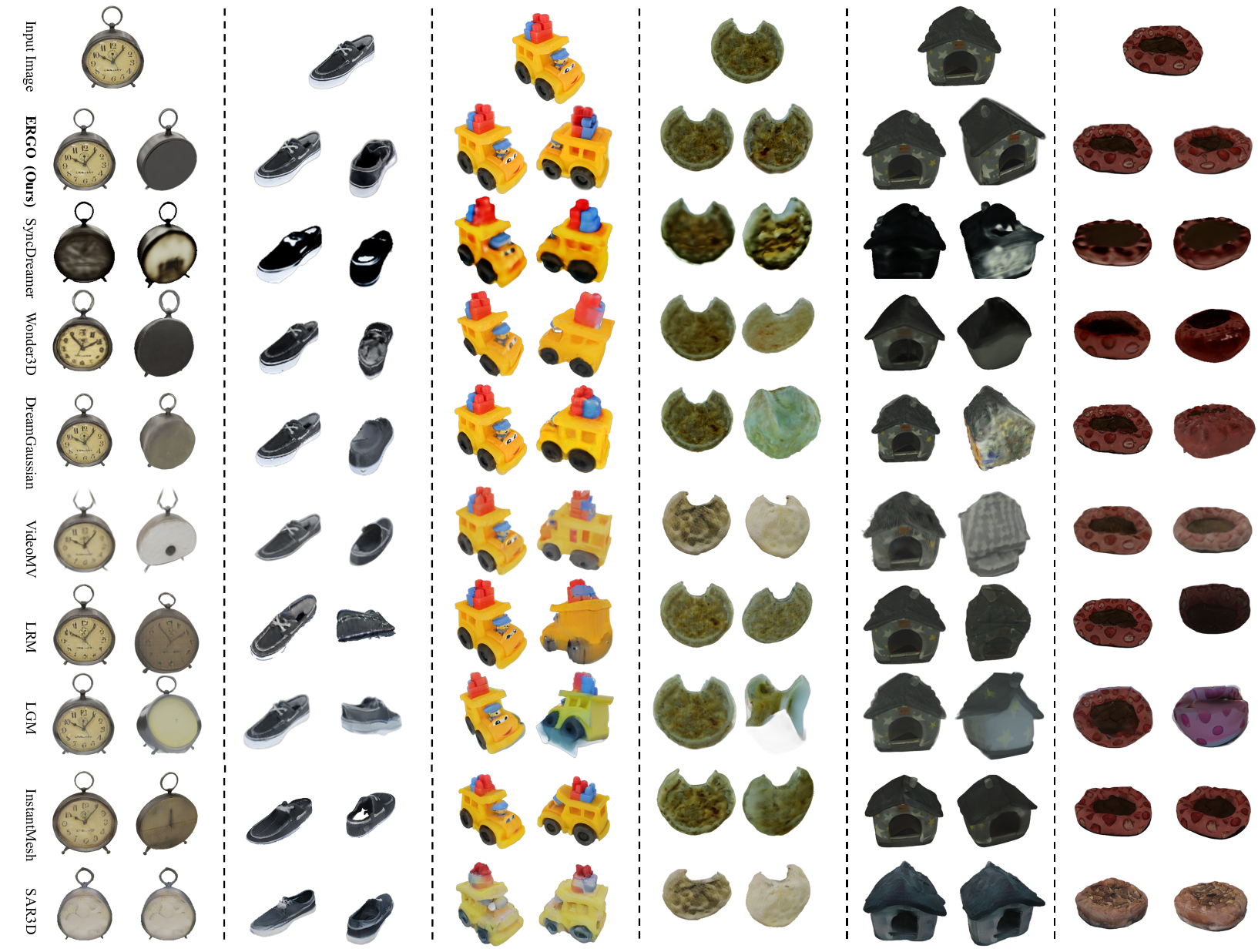}
  \caption{Qualitative comparison on the GSO dataset~\cite{downs2022google} and the OmniObject3D dataset~\cite{wu2023omniobject3d}.}
  \label{fig: qulitative_benchmark}
\end{figure*}

\begin{figure*}[!t]
  \centering
  \includegraphics[width=1.0\hsize]{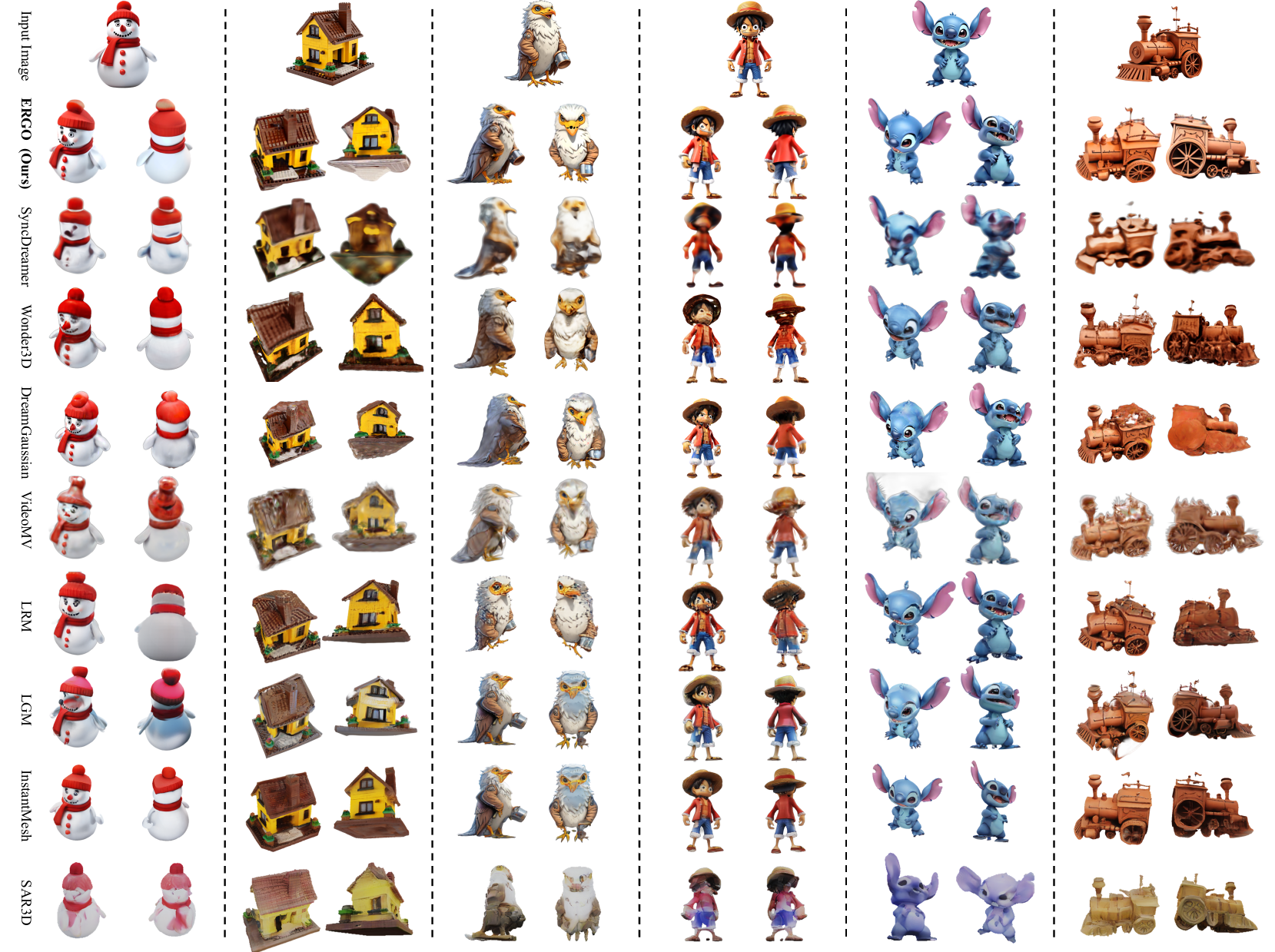}
  \caption{Qualitative comparisons on in-the-wild images.}
  \label{fig: qulitative_in_the_wild}
\end{figure*}

\subsection{Setup}

\noindent\textbf{Implementation details}. 
All experiments are conducted on a single NVIDIA A100 GPU with 40 GB GPU memory. We adopt Zero123++ \cite{shi2023zero123plus} as our base MVD model. For each test case, we randomly initialize 5000 3D Gaussian particles within a sphere of radius 0.5. Each particle is set with an initial opacity value of 0.1 and a grey color with [r, g, b]=[128, 128, 128]. The Gaussian particles are optimized for 1500 interactions using the Adam optimizer~\cite{KingBa15}. The learning rate for the position attribute is initialized to $1 \times 10^{-3}$ and is gradually decayed to $2 \times 10^{-5}$. For the remaining Gaussian attributes, including scale, rotation, opacity, and color, the learning rates are set to $5 \times 10^{-3}$, $5 \times 10^{-3}$, $5 \times 10^{-2}$, and $1 \times 10^{-2}$, respectively. 3D Gaussian densification is carried out every 100 iterations, while the opacity reset operation is performed every 500 iterations. Following \cite{yang2024gaussianobject}, we remove floaters every 400 iterations using the K-Nearest Neighbors algorithm. For the rendering of auxiliary images, the azimuth angle is sampled within the range $\left [-180^{\circ}, 180^{\circ} \right]$, and the elevation angle is constrained to $\left [-30^{\circ}, 30^{\circ} \right ]$. The rendering resolution increases incrementally from $128 \times 128$ to $512 \times 512$ when computing the SDS loss, while remaining constant at $320 \times 320$ for pixel-wise loss calculations with a fixed radius of 2 and FOV of $49.1^{\circ}$. The hyperparameters for balancing objectives are set as $\lambda_{v} = 1 \times 10^4$, $\lambda_{d}=10$, $\lambda_{m}=1 \times 10^3$, and $\lambda_{t}=1$. The scaling factor $s$ for the warping process in the geometry objective is set to 4, and the sensitivity control of weight updates $\eta$ is set to 3. 

\noindent\textbf{Datasets}. 
We utilize two publicly available 3D datasets, i.e., the Google Scanned Objects (GSO) dataset~\cite{downs2022google} and the OmniObject3D dataset~\cite{wu2023omniobject3d}, both of which contain a diverse range of object categories. On the GSO dataset, we utilize the same 30 objects as used in SyncDreamer~\cite{liu2024syncdreamer}, spanning everyday items to animals. Furthermore, we randomly select 50 objects from OmniObject3D, ensuring a wide variety of categories to verify the versatility and effectiveness of the proposed method. For each object, a single image with a resolution of $320 \times 320$ is rendered as the reference image, while 16 multi-view images with camera poses uniformly distributed over an azimuth range of $0^{\circ}$ to $360^{\circ}$ and with a fixed elevation angle of $-30^{\circ}$.

\begin{table*}[!t]
    \caption{Ablation study on the proposed components. GAO, TAO, and AW represent geometry-aware objective, texture-aware objective, and adaptive weights, respectively.
    \label{tab: ablation_results}}
    \normalsize
    \def\arraystretch{1.1}
    \footnotesize
    \tabcolsep 6.0pt
    \centering
    \begin{tabular}{c c c c c c c c c c c c}
        \toprule
        \multirow{2}{*}{Method} & \multirow{2}{*}{GAO} & \multirow{2}{*}{TAO} & \multirow{2}{*}{AW} & & \multicolumn{3}{c}{$\left| \text{azimuth} \right| \le 90^\circ$} & & \multicolumn{3}{c}{$\left| \text{azimuth} \right| > 90^\circ$} \\
        \cmidrule{6-8} \cmidrule{10-12}
        & & & & & PSNR$\uparrow$ & SSIM$\uparrow$ & LPIPS$\downarrow$ & & PSNR$\uparrow$ & SSIM$\uparrow$ & LPIPS$\downarrow$ \\
        \midrule
        Baseline & \xmark & \xmark & \xmark & & 21.97 & 0.8541 & 0.1458 & & 18.79 & 0.8214 & 0.1838 \\
        Baseline + GAO  & \cmark & \xmark & \xmark & & 22.54 & 0.8613 & 0.1287 & & 19.32 & 0.8261 & 0.1689 \\
        Baseline + TAO & \xmark & \cmark & \xmark & & 22.12 & 0.8582 & 0.1422 & & 19.00 & 0.8258 & 0.1822 \\
        ERGO w/o AW & \cmark & \cmark & \xmark & & 22.68 & 0.8690 & 0.1282 & & 19.42 & 0.8327 & 0.1616 \\
        \midrule
        ERGO & \cmark & \cmark & \cmark & & \textbf{22.94} & \textbf{0.8813} & \textbf{0.1220} & & \textbf{19.62} & \textbf{0.8397} & \textbf{0.1590} \\
        \bottomrule
    \end{tabular}
\end{table*}

\noindent\textbf{Baselines and evaluation metrics}.
We consider both optimization-based methods and feed-forward large reconstruction models for comparison. Specifically, we select three representative optimization-based methods, including SyncDreamer~\cite{liu2024syncdreamer}, Wonder3D~\cite{long2024wonder3d}, and DreamGaussian~\cite{tang2024dreamgaussian}. Meanwhile, five cutting-edge large reconstruction models are selected, including LRM~\cite{openlrm}, LGM~\cite{tang2024lgm}, VideoMV \cite{zuo2024videomv}, InstantMesh \cite{xu2024instantmesh}, and SAR3D \cite{chen2024sar3d}. All these baselines are evaluated with their provided code and models. To conduct a multi-faceted assessment of generation quality, three widely used metrics, including PSNR~\cite{hore2010image}, SSIM~\cite{1284395}, and LPIPS~\cite{8578166}, are used for evaluation from the aspects of texture quality, geometry quality, and perceptual quality, respectively.

\begin{figure*}[!t]
  \centering
  \includegraphics[width=1.0\hsize]{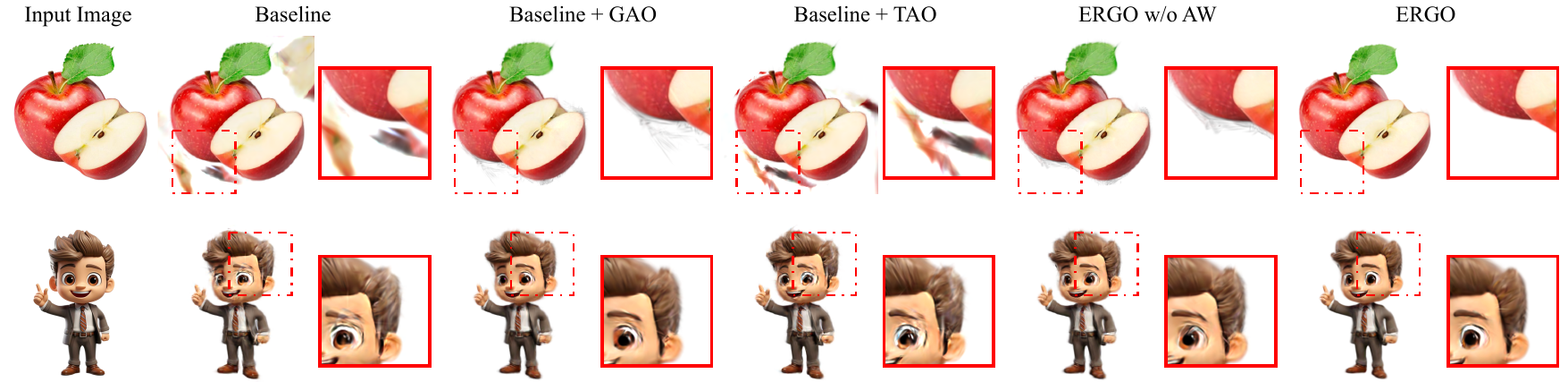}
  \caption{Visual examples of the ablation study on the three key components of ERGO, namely, the geometry-aware objective (GAO), the texture-aware objective (TAO), and the adaptive weighting mechanism (AW).}
  \label{fig: vis_ablation}
\end{figure*}

\subsection{Comparison with State-of-the-art Methods}

\textbf{Quantitative comparisons}. The quantitative performance of ERGO against the eight selected state-of-the-art methods on GSO and OminiObject3D is reported in Table \ref{tab:quantitative_results}. ERGO consistently achieves the best performance across all three evaluation metrics on both datasets. Particularly, our implementation of 3DGS is built upon DreamGaussian \cite{tang2024dreamgaussian}, and the performance gains of ERGO are significant, e.g., the SSIM values of ERGO on GSO and OminiObject3D are $0.8426$ and $0.8854$, while those of DreamGaussian are $0.8048$ and $0.8072$. Furthermore, ERGO obtains the highest PSNR and the lowest LPIPS, indicating that its generated 3D content has higher texture quality and cross-view consistency. Among the competing methods, VideoMV~\cite{zuo2024videomv} achieves the closest performance to ERGO, which we attribute to its exploitation of video priors that inherently encode temporal smoothness as a proxy for multi-view geometric and textural consistency. Nevertheless, ERGO surpasses VideoMV by employing the geometry-aware objective and the texture-aware objective with adaptive weights, thereby achieving more robust optimization under inconsistent multi-view supervision.

\noindent\textbf{Qualitative comparisons on GSO and OminiObject3D}. Figure \ref{fig: qulitative_benchmark} shows the qualitative results of ERGO and the baselines. One can see that SyncDreamer \cite{liu2024syncdreamer} tends to generate smooth geometry but often lacks finer details. While Wonder3D~\cite{long2024wonder3d} captures more detailed textures, it sometimes produces unrealistic geometry due to inconsistencies in sparse multi-view images. DreamGaussian~\cite{tang2024dreamgaussian} can generate 3D models that match the reference image, but it tends to produce blurry textures in unseen areas, as it relies solely on SDS optimization to supplement novel view information. As for feed-forward large models,
LRM~\cite{openlrm} is prone to generating grid artifacts.
LGM~\cite{tang2024lgm}, which integrates the multi-view diffusion model ImageDream~\cite{wang2023imagedream}, is limited by the quality of the generated multi-views and often produces artifacts in unseen areas. Similar artifacts also exhibit in the results of other methods like VideoMV \cite{zuo2024videomv}, InstanceMesh \cite{xu2024instantmesh}, and SAR3D \cite{chen2024sar3d}. In contrast, our ERGO framework generates high-quality 3D Gaussian models that not only maintain consistency across various camera poses but also excel in producing realistic textures with richer details, demonstrating its superiority over the baseline methods.

\begin{figure*}[!t]
  \centering
  \includegraphics[width=1.0\hsize]{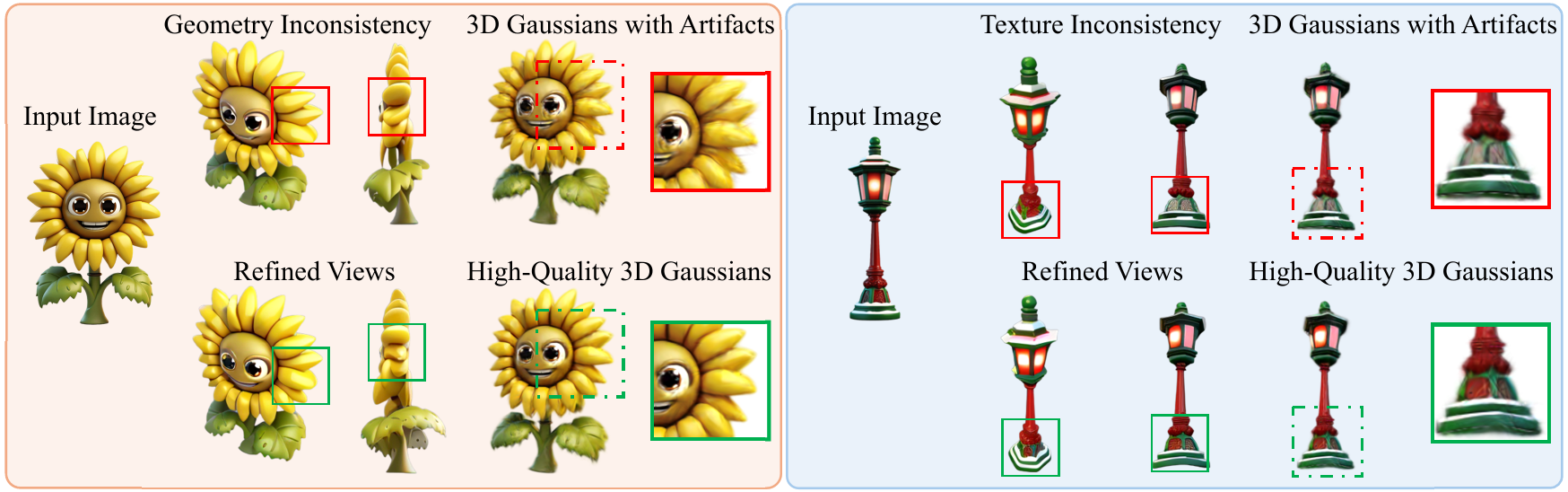}
  \caption{Visual examples of artifacts produced by performing 3DGS on inconsistent multi-view images (labeled in red). These artifacts are mitigated by ERGO (labeled in green).}
  \label{fig: ddim_improve}
\end{figure*}

\noindent\textbf{Qualitative comparisons on in-the-wild images}. To evaluate the generalization ability of ERGO, we test ERGO on diverse in-the-wild images, and the results are shown in Figure \ref{fig: qulitative_in_the_wild}. ERGO demonstrates robust performance on objects with complex geometries and intricate textures, effectively mitigating artifacts observed in existing methods, such as over-smoothing textures (e.g., the grids of the house in the second column), cross-view inconsistency (the red hats in the first column), and distorted geometric structures (the wheels in the last column). These results validate that the adaptive weighting mechanism of ERGO successfully suppresses supervision noise while preserving high-frequency details and geometric integrity in challenging real-world scenarios.

\subsection{Ablation Studies}

\textbf{Effects of the Proposed Components}. We conduct an ablation study to thoroughly evaluate the efficacy of the proposed components in ERGO. In this experiment, 3D items are randomly sampled from the GSO dataset~\cite{downs2022google}. Starting with an azimuth of 0 degrees, we render 16 views uniformly across the range of [-180, 180] degrees. We construct three variants of ERGO, including the baseline with the geometry-aware objective, the baseline with the texture-aware objective, and ERGO without adaptive weights, to evaluate the three key components of ERGO. From the results in Table \ref{tab: ablation_results}, we can see that these three components improve the performance of the baseline consistently. The performance gains of the baseline with the geometry-aware objective and those with the texture-aware objective are similar, indicating that both geometry and texture guidance are important for the image-to-3D task. The results of the full framework against those of the variant without adaptive weightings validate that exploiting excess risk to modulate the optimization process yields better performance. Figure \ref{fig: vis_ablation} compares the results of ERGO and its variants. These visual examples provide an intuitive demonstration of the efficacy of the proposed components. For instance, in the first row, we can see that the baseline generates blurry floaters around its boundaries. These floaters are removed significantly by the geometry-aware objective, while they remain but are sharper with the texture-aware objective. Finally, unifying these components into the ERGO framework obtains the most balanced and compelling results. 

\noindent\textbf{Effects of View Variations}. One of our assumptions is that reconstruction quality degrades as the angular separation between a rendering view and the input reference view increases. To validate this, we partition evaluation views into two groups based on absolute azimuth deviation, i.e., $\left | \text{azimuth} \right | \le 90$ and $\left | \text{azimuth} \right | > 90$. The experimental results are reported in Table \ref{tab: ablation_results}. These results confirm a strong correlation between angular proximity to the reference view and reconstruction fidelity. The performance of both ERGO and the baseline under $\left | \text{azimuth} \right | \le 90$ is consistently superior to that under $\left | \text{azimuth} \right | \le 90$. Nevertheless, ERGO obtains considerable performance gains in both groups, validating that our adaptive weighting mechanism effectively suppresses unreliable supervision signals from geometrically distant views.  

\noindent\textbf{Effects of Multi-view Inconsistency}. In addition to the above visual examples in the ablation study of the proposed components, we demonstrate the effects of using multi-view images with geometry and texture inconsistency for 3D creation in Figure \ref{fig: ddim_improve}. We can see that employing 3DGS directly on inconsistent images results in blurry and erroneous textures. These artifacts are mitigated through two mechanisms in ERGO. First, the geometry correction step that inverts renderings from 3D Gaussians back to the base MVD model refines multi-view textures, e.g, the bottom of the lamp in the refined views in Figure \ref{fig: ddim_improve}. Second, the adaptive weighted objectives of ERGO further enhance the geometry structures and textures of the created 3D objects, as the effects of inconsistent image regions are suppressed. 

\section{Conclusion}
In this paper, we introduce ERGO, an excess-risk-guided optimization framework for single-image 3D reconstruction. ERGO constructs a global-local adaptive paradigm to tackle the critical geometric and textural inconsistency issues and enable seamless integration of multi-view diffusion models with optimization-based methods. From the global perspective, an excess-risk-derived objective weighting mechanism is introduced, while from the local perspective, a geometry-aware objective and a texture-aware objective are designed to achieve regional modulation. Extensive experiments on two public datasets as well as in-the-wild images validate that ERGO effectively mitigates spurious supervision signals and texture blurring artifacts, significantly enhancing the cross-view consistency and fine-grained fidelity of the reconstructed 3D content.  
{
    \small
    \bibliographystyle{ieeenat_fullname}
    \bibliography{main}
}


\end{document}